\documentclass[11pt]{article}

\usepackage[]{acl}
\usepackage{times}
\usepackage{latexsym}
\usepackage[T1]{fontenc}
\usepackage[utf8]{inputenc}
\usepackage{microtype}
\usepackage{inconsolata}
\usepackage{graphicx}
\usepackage{amsmath}
\usepackage{amssymb}
\usepackage{booktabs}
\usepackage{placeins}
\usepackage{float}
\usepackage{caption}

\title{Interpretable Semantic Gradients in SSD: A PCA Sweep Approach and a Case Study on AI Discourse}

\author{
  Hubert Plisiecki \\
  IDEAS Research Institute \\
  \texttt{hplisiecki@gmail.com}
  \And
  Maria Leniarska \\
  VIZJA University \\
  \AND
  Jan Piotrowski \\
  Warsaw University of Technology \\
  \And
  Marcin Zajenkowski \\
  University of Warsaw \\
}

\begin{document}
\maketitle

\begin{abstract}
Supervised Semantic Differential (SSD) is a mixed quantitative–interpretive method that models how text meaning varies with continuous individual-difference variables by estimating a semantic gradient in an embedding space and interpreting its poles through clustering and text retrieval. SSD applies PCA before regression, but currently no systematic method exists for choosing the number of retained components, introducing avoidable researcher degrees of freedom in the analysis pipeline. We propose a \emph{PCA sweep} procedure that treats dimensionality selection as a joint criterion over representation capacity, gradient interpretability, and stability across nearby values of~$K$. We illustrate the method on a corpus of short posts about artificial intelligence written by Prolific participants who also completed Admiration and Rivalry narcissism scales. The sweep yields a stable, interpretable Admiration-related gradient contrasting optimistic, collaborative framings of AI with distrustful and derisive discourse, while no robust alignment emerges for Rivalry. We also show that a counterfactual using a high-PCA dimension solution heuristic produces diffuse, weakly structured clusters instead, reinforcing the value of the sweep-based choice of~$K$. The case study shows how the PCA sweep constrains researcher degrees of freedom while preserving SSD’s interpretive aims, supporting transparent and psychologically meaningful analyses of connotative meaning.

\end{abstract}

\section{Introduction}

Supervised Semantic Differential (SSD) is a recent mixed-method that analyzes shifts in meaning of concepts and texts with regards to psychological individual differences of their authors \citep{plisiecki_measuring_2025}. The method is inspired by longstanding psychological Semantic Differential methods that measure connotative meaning of concepts via Likert-scales with polar concept opposites (e.g. warm/cold, strong/weak; \citealp{osgood_measurement_1978}) and by modern distributional semantics, where word embeddings model relational structure in language use \citep{mikolov_efficient_2013, kozlowski_geometry_2019, garg_word_2018}. Given texts annotated with an outcome variable, such as a trait score or attitude scale, SSD creates individual representations of specific concepts, also called Personal Concept Vectors (PCVs) by aggregating words that collocate with a lexicon of interest relating to a concept under analysis. SSD then fits a linear model in a reduced embedding space with regards to the outcome variable and uses the regression coefficients as a semantic direction or gradient, whose positive and negative poles are explored through clustering of nearby words and further elucidated by retrieving original text snippets which numerical representations are the most aligned to each cluster centroid. The statistical results of the regression are analyzed to estimate whether a linear shift in individual semantic representation explains a significant portion of the variance in the outcome variable.

In the original SSD method, the PCVs are compressed with principal component analysis (PCA) before regression in order to make it possible to apply it to small corpora, and reduce the numer of redudant dimensions \citep{jolliffe_principal_2002}. However, as of now there is no systematic method of choosing the number of components (K) to be extracted. While original results were relatively stable across similar component ranges, leaving the choice of dimensionality to the researcher introduces an avoidable researcher degree of freedom, thereby increasing the risk of overfitting, reducing the transparency of the analysis pipeline, and potentially biasing substantive interpretations of the resulting semantic gradients \citep{simmons_false-positive_2011}. 

This work proposes a simple computational heuristic that can be used to pick K in a principled way. We introduce a PCA sweep procedure that treats K selection as a joint optimization problem over topic representation quality, interpretability, and the stability of the results. The sweep evaluates a sequence of K values, fits SSD at each K and tracks both cluster coherence (used as an interpretability criterion; \citealp{mimno_optimizing_2011}), their allignment with the gradient, as well as a result  stability measure based on cosine differences between consecutive gradients.

We showcase the applicability of SSD with a PCA sweep by applying it to a dataset of posts written by Prolific study participants who have also answered a questionnaire assessing their Admiration (ADM) and Rivalry (RIV) Narcissistic traits \citep{back_narcissistic_2018}. The case study serves primarily as an illustration of the extended method; the main contribution is methodological.

\section{Background: Supervised Semantic Differential}

Supervised Semantic Differential (SSD) assumes a collection of documents $d_i$ paired with continuous outcomes $y_i$, such as trait or attitude scores. In its canonical formulation \citep{plisiecki_measuring_2025}, SSD focuses on a specific concept of interest by means of a lexicon: words that denote or closely relate to the target concept (e.g., \emph{abortion}, \emph{immigration}). These lexicon terms are used to identify and aggregate the local semantic contexts in which the concept appears, yielding a PCV for each author, reflecting how the concept is represented in their texts.

Each document is mapped to a dense vector $\mathbf{x}_i \in \mathbb{R}^D$ using a fixed embedding model and a composition scheme (e.g., SIF-weighted averages with removal of the top principal component to reduce anisotropy; \citealp{arora_simple_2017, mu_all-but--top_2017}). To reduce redundancy and to make the regression applicable in small datasets, SSD applies PCA to the vectors and projects them to a lower-dimensional representation $\tilde{\mathbf{x}}_i \in \mathbb{R}^K$. A linear model is then estimated:
\[
y_i = \alpha + \boldsymbol{\beta}^\top \tilde{\mathbf{x}}_i + \epsilon_i.
\]
The regression coefficients are normalized to unit length to obtain a semantic gradient $\hat{\boldsymbol{\beta}}$, which is back-projected to the original embedding space. SSD explains the positive and negative poles of this gradient by retrieving nearest neighbors, clustering them separately on each side, and retrieving original text snippets aligned with cluster centroids. These clusters provide qualitative evidence of how the meaning of the focal concept shifts as the outcome increases.

The choice of PCA dimensionality $K$ affects both regression and interpretation: too few components may collapse distinct semantic regularities, whereas too many may introduce unstable or noisy directions. A principled way to choose $K$ is therefore essential.

\section{PCA Sweep for SSD}

We introduce a \emph{PCA sweep} procedure that operationalizes the choice of $K$ as a structured model-selection problem. The key principle is that $K$ is not optimized for predictive accuracy, but for three properties central to SSD as an interpretive method: (i) the quality of the semantic representation after dimensionality reduction, (ii) the interpretability of the derived semantic gradient, and (iii) the stability of that gradient across nearby values of~$K$. Although these diagnostics are computed after the regression is fitted, they are not functions of variance explained by the regression or predictive fit, and therefore do not reward models that align more strongly with the outcome variable. Instead, they privilege solutions whose semantic structure is coherent and stable across $K$.

Given a user-specified range of candidate dimensionalities (e.g., $K\in\{20,22,\dots,120\}$), the sweep runs SSD repeatedly with identical preprocessing, weighting, and clustering settings, and records diagnostic quantities for each~$K$.

\textbf{Representation statistics.} For every configuration, we compute the cumulative proportion of variance explained by the first $K$ principal components. This provides a monotonic measure of representation capacity: larger $K$ values encode more information from the PCVs but also increase the risk of capturing idiosyncratic or noisy variation. Because this quantity grows mechanically with $K$, it is used only as a baseline for detrending rather than as a selection criterion.

\textbf{Interpretability diagnostics.} After fitting the regression and back-projecting the semantic gradient $\hat{\boldsymbol{\beta}}_K$ to the embedding space, SSD retrieves nearest neighbors on the positive and negative poles and clusters them separately. For each cluster we compute internal coherence and the cosine alignment between its centroid and $\hat{\boldsymbol{\beta}}_K$. These are aggregated into a single, cluster size-weighted \emph{interpretability score} that summarizes how well the learned direction organizes semantically meaningful neighborhoods. To account for the fact that higher-dimensional representations tend to yield higher scores even when the underlying structure does not qualitatively improve, the aggregate is detrended with respect to the (log-transformed) variance explained, and the residuals are standardized to $z$-scores.

\textbf{Stability diagnostics.} The sweep also tracks how the semantic gradient changes as $K$ increases. Let $\hat{\boldsymbol{\beta}}_{K}$ and $\hat{\boldsymbol{\beta}}_{K-1}$ be consecutive normalized gradients (back-projected to the original space). We measure \emph{gradient change}
\[
\Delta_K = 1 - \cos\!\left(\hat{\boldsymbol{\beta}}_{K},\,\hat{\boldsymbol{\beta}}_{K-1}\right),
\]
where lower values indicate greater stability. Intuitively, highly unstable regimes signal that the semantic structure depends on components that are weakly supported by the data.

\textbf{Plateau-sensitive smoothing.} Because the interpretability is highly volatile at low K, instead of rewarding isolated local maxima, both the detrended interpretability curve and the stability curve are smoothed using a local neighborhood average (AUCK), which emphasizes broad, stable plateaus over sharp spikes. This favors configurations where interpretability remains high while the gradient has largely converged.

\textbf{Selection rule.} For each $K$, the sweep computes a joint score combining interpretability and stability:
\[
\text{joint\_score}_K = \tfrac{1}{2}\bigl(\text{interp\_auck}_K + \text{stab\_auck}_K\bigr).
\]
The selected dimensionality is the \emph{smallest} $K$ attaining the maximal joint score, privileging parsimonious, stable, and interpretable solutions.

\section{Case Study: Admiration and Rivalry in AI Discourse}
\label{sec:case-study}

We applied SSD with the PCA sweep to a corpus of $N{=}349$ short posts about artificial intelligence (AI) written by Prolific participants (mean length $\approx 30$ words; full writing prompt and additional study details provided in Appendix~\ref{app:prompt}), who also completed Admiration (ADM) and Rivalry (RIV) narcissism scales reflecting the two self-regulatory strategies in the Narcissistic Admiration and Rivalry Concept \citep{back_narcissistic_2013}, where Admiration captures assertive self-enhancement and status-seeking tendencies, and Rivalry reflects defensive, antagonistic self-protection in response to perceived threat \citep{back_narcissistic_2018}. Tokenization and linguistic preprocessing were performed using spaCy~3.8.7 with the \texttt{en\_core\_web\_sm} English model~\citep{ines_montani_explosionspacy_2023}. Posts were embedded using the 300 dimensional Dolma GloVe model \citep{carlson_new_2025} with SIF weighting ($a{=}10^{-3}$), removal of the top principal component. Crucially, a lexicon was not used to dial down on AI-specific terms, because the posts did not rely on a consistent set of shared AI keywords; however, since all texts were produced under the same AI-focused prompt, each post can be treated as a whole as a meaningful reflection of participants’ reactions to AI.  The sweep evaluated $K\!\in\!\{1,3,\dots,119\}$ with cluster number chosen by silhouette from ranges $k\!\in\![2,5]$, top-100 neighbors per pole, AUCK radius $=3$, and median-smoothed gradient change (win$=7$); remaining settings followed the original SSD configuration (see associated code).

Table~\ref{tab:ai-regression} summarizes the regression results at the sweep-selected $K$ values. For ADM, the model explained a small–moderate but reliable proportion of variance ($R^2_{\text{adj}}=.19$, $F=6.32$, $p<10^{-10}$, $r\approx.47$) with a pronounced semantic gradient ($\|\hat{\beta}\|=5.58$; $\Delta_{0.1}=0.65$). By contrast, the RIV model did not reach significance ($R^2_{\text{adj}}=.03$, $p=.095$), and we therefore restrict interpretation to ADM.

\begin{center}
\small
\begin{tabular}{lrrrrrr}
\toprule
Trait & $K$ & $R^2_{\text{adj}}$ & $F$ & $p$ & $r$ & $\|\hat{\beta}\|$ \\
\midrule
ADM & 15 & .19 & 6.32 & $<\!10^{-10}$ & .47 & 5.58 \\
RIV & 23 & .03 & 1.43 & .095 & .30 & 5.38 \\
\bottomrule
\end{tabular}

\captionof{table}{SSD regression results for AI posts at the sweep-selected $K$.}
\label{tab:ai-regression}
\end{center}

\begin{center}
\includegraphics[width=\linewidth]{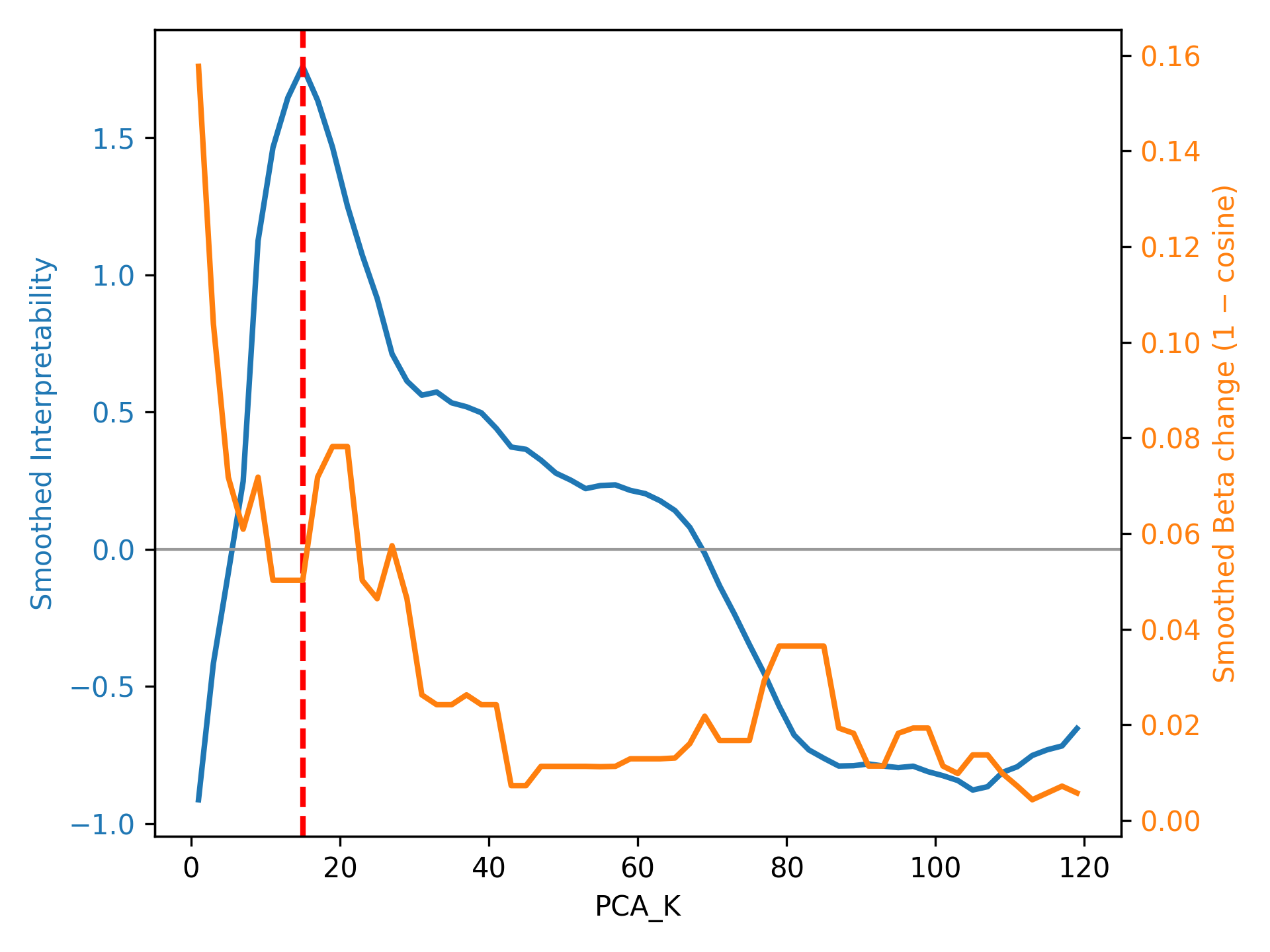}

\captionof{figure}{Sweep diagnostics for ADM: detrended interpretability and smoothed unit-change curves, with the selected $K{=}15$ plateau.}
\label{fig:interpretability}
\end{center}

Figure~\ref{fig:interpretability} presents the sweep diagnostics for ADM. The detrended interpretability curve rises sharply at low $K$ and reaches a peak around $K{=}15$ (explaining 50\% of variance in the original PCVs). The unit-change curve exhibits a local plateau at the same position; although it declines for larger $K$, this drop is mirrored by a decrease in interpretability. Taken together, these trends indicate that $K{=}15$ lies in the most parsimonious region where the gradient remains both interpretable and stable.

To interpret the ADM gradient, we clustered nearest neighbours on the positive and negative poles. Table~\ref{tab:ai-clusters} summarizes the resulting themes with snippets of relevant representative excerpts.

\begin{center}
\small
\begin{tabular}{ccp{0.68\linewidth}}
\toprule
Pole & Size & Summary (Top Words / Excerpt) \\
\midrule
$+$ & 14 & \emph{Cultivation \& enrichment}:
\textit{cultivate, rediscover, rejuvenated} — ``AI's potential is boundless\dots'' \\
$+$ & 86 & \emph{Innovation \& collaboration}:
\textit{innovation, partnership, empower} — ``AI is transforming our world\dots'' \\
\midrule
$-$ & 56 & \emph{Deception \& threat}:
\textit{misleading, dishonest, unfair} — ``woke, evasive and bizarre responses\dots'' \\
$-$ & 44 & \emph{Ridicule \& contempt}:
\textit{ridiculous, absurd, laughable} — ``the stupid programmers exposed their hand'' \\
\bottomrule
\end{tabular}

\captionof{table}{Positive and negative ADM clusters in AI discourse.}
\label{tab:ai-clusters}
\end{center}

The positive pole reflects a \emph{future-oriented, collaborative, and prosocial framing of AI}, linking it to innovation, integration, and constructive technological progress. The negative pole instead emphasizes \emph{distrust, antagonism, and derision}, portraying AI and its creators as deceptive, biased, or ideologically motivated. Overall, the case study illustrates how the PCA sweep supports interpretable SSD analysis by constraining dimensionality in a transparent, stability-aware manner.

For comparison, Appendix~\ref{app:high-dim} reports a counterfactual SSD run at an arbitrarily high dimensionality ($K{=}120$), where the resulting clusters become diffuse and hard to interpret meaningfully.

\section{Discussion and Conclusion}

This work introduced a PCA sweep procedure for SSD that selects dimensionality by jointly prioritizing gradient stability and semantic interpretability rather than predictive fit. In the AI case study, the sweep yielded a compact representation with coherent cluster structure and stable gradients, whereas the counterfactual high-dimensional run in Appendix~\ref{app:high-dim} produced diffuse and weakly structured clusters. This contrast demonstrates that the sweep not only constrains researcher degrees of freedom, but also helps preserve the qualitative aims of SSD by discouraging over-parameterized representations that cease to reflect psychologically meaningful structure.

Beyond the methodological contribution, the case study revealed a psychologically interpretable association between Admiration and the semantics of AI discourse. The positive pole of the ADM gradient reflected future-oriented, collaborative, and prosocial framings of AI, while the negative pole emphasized distrust, antagonism, and ridicule. This contrast is consistent with accounts of Admiration as an agentic, self-enhancing orientation toward status and positively valenced self-views \citep{back_narcissistic_2018,zeiglerhill_narcissism_2019}, suggesting that individuals higher in Admiration may align themselves with culturally valued narratives of technological progress and collective advancement. A complementary interpretation draws on prior work describing large AI systems as compliant and \emph{sycophantic} in evaluative interaction \citep{perez_discovering_2022}; semantics that frame AI as supportive, empowering, or socially harmonious may resonate more strongly with affirmation-oriented self-regulatory motives linked to Admiration.

More broadly, the results illustrate how SSD, when paired with a principled choice of PCA dimensionality, can surface stable, interpretable semantic gradients that connect language use to underlying psychological dispositions. However, the PCA sweep addresses only one source of flexibility in the SSD pipeline. A key open challenge is developing similarly principled criteria for upstream modeling choices - most notably, the selection of the base embedding model. Different embedding spaces encode distinct cultural, temporal, and stylistic regularities, and these choices can meaningfully shape the structure of recovered gradients. Extending the logic of stability- and interpretability-based diagnostics to the level of model selection represents an important next step toward a fully transparent and methodologically grounded SSD workflow.

Overall, the sweep procedure strengthens SSD as a mixed quantitative–interpretive method by tying dimensionality selection directly to gradient stability and semantic coherence. The approach offers a lightweight, model-agnostic criterion that can be integrated into embedding-based analysis in order to both enhance interpretability and reduce researcher degrees of freedom.

\section*{Limitations}

Our case study is based on a relatively small, survey-elicited dataset of short AI-related posts from Prolific participants, which limits the generalizability of the observed semantic gradients beyond this population and task context. The analysis also operates at the level of whole-text representations rather than concept-specific lexicon-based PCVs; while justified by the uniform prompting, this choice may blur finer-grained semantic distinctions. Moreover, the PCA sweep addresses only one source of flexibility in SSD, namely the dimensionality selection, while other design choices remain open, such as the choice of the embedding model, and word window size when using a lexicon. As the appendix counterfactual illustrates, different representational assumptions can meaningfully affect interpretability, and a broader framework for principled model and parameter selection in SSD remains an open methodological challenge. Finally, the psychological interpretations offered here are correlational and depend on the alignment between semantic structure and trait scores; future work should test the robustness of these gradients across datasets, languages, and outcomes, as well as test the case study conclusions with targeted laboratory studies. 

\section*{Ethical considerations}

The present work analyzes text data paired with individual-difference measures, raising issues related to privacy, interpretative uncertainty, and the potential misuse of trait–language associations. Importantly, SSD is not designed as a predictive or profiling technology: the method estimates weak, low-variance semantic gradients that support qualitative interpretation rather than accurate inference about individuals. As such, the approach is poorly suited for identifying, classifying, or predicting participants’ psychological dispositions, and we explicitly discourage such uses. Our analyses are conducted at the group and pattern level, and all interpretations are contingent, correlational, and theoretically guided rather than diagnostic. More broadly, we view SSD as a tool for hypothesis generation and meaning exploration in psychological and social research, not as a mechanism for automated assessment or decision making. The data was confirmed to not include any personally identifying or offensive content. The text of this manuscript was partially polished with the assistance of a Large Language Model; all automatically  revised passages were reviewed, validated, and, where necessary, corrected by the authors to ensure accuracy, faithfulness to the data, and conceptual integrity. Future applications of SSD should continue to prioritize transparency, participant respect, and careful communication of uncertainty.
\bibliography{references}

@book{osgood_measurement_1978,
	address = {Urbana-Champaign},
	title = {The measurement of meaning},
	isbn = {978-0-252-74539-3},
	language = {eng},
	publisher = {University of Illinois Press},
	author = {Osgood, Charles Egerton and Suci, George J. and Tannenbaum, Percy H.},
	year = {1978},
}

@article{kozlowski_geometry_2019,
	title = {The {Geometry} of {Culture}: {Analyzing} the {Meanings} of {Class} through {Word} {Embeddings}},
	volume = {84},
	issn = {0003-1224, 1939-8271},
	shorttitle = {The {Geometry} of {Culture}},
	url = {https://journals.sagepub.com/doi/10.1177/0003122419877135},
	doi = {10.1177/0003122419877135},
	language = {en},
	number = {5},
	urldate = {2026-01-02},
	journal = {American Sociological Review},
	author = {Kozlowski, Austin C. and Taddy, Matt and Evans, James A.},
	month = oct,
	year = {2019},
	pages = {905--949},
}

@inproceedings{arora_simple_2017,
	title = {A {Simple} but {Tough}-to-{Beat} {Baseline} for {Sentence} {Embeddings}},
	url = {https://openreview.net/forum?id=SyK00v5xx},
	language = {en},
	urldate = {2026-01-02},
	author = {Arora, Sanjeev and Liang, Yingyu and Ma, Tengyu},
	month = feb,
	year = {2017},
}

@book{jolliffe_principal_2002,
	address = {New York},
	series = {Springer {Series} in {Statistics}},
	title = {Principal {Component} {Analysis}},
	copyright = {http://www.springer.com/tdm},
	isbn = {978-0-387-95442-4},
	url = {http://link.springer.com/10.1007/b98835},
	language = {en},
	urldate = {2026-01-02},
	publisher = {Springer-Verlag},
	author = {Jolliffe, I. T.},
	year = {2002},
	doi = {10.1007/b98835},
}

@misc{mu_all-but--top_2017,
	title = {All-but-the-{Top}: {Simple} and {Effective} {Postprocessing} for {Word} {Representations}},
	copyright = {arXiv.org perpetual, non-exclusive license},
	shorttitle = {All-but-the-{Top}},
	url = {https://arxiv.org/abs/1702.01417},
	doi = {10.48550/ARXIV.1702.01417},
	urldate = {2026-01-02},
	publisher = {arXiv},
	author = {Mu, Jiaqi and Bhat, Suma and Viswanath, Pramod},
	year = {2017},
	note = {Version Number: 2},
}

@incollection{back_narcissistic_2018,
	address = {Cham},
	title = {The {Narcissistic} {Admiration} and {Rivalry} {Concept}},
	isbn = {978-3-319-92171-6},
	url = {https://doi.org/10.1007/978-3-319-92171-6_6},
	language = {en},
	urldate = {2026-01-02},
	booktitle = {Handbook of {Trait} {Narcissism}: {Key} {Advances}, {Research} {Methods}, and {Controversies}},
	publisher = {Springer International Publishing},
	author = {Back, Mitja D.},
	editor = {Hermann, Anthony D. and Brunell, Amy B. and Foster, Joshua D.},
	year = {2018},
	doi = {10.1007/978-3-319-92171-6_6},
	pages = {57--67},
}

@misc{plisiecki_measuring_2025,
	title = {Measuring {Individual} {Differences} in {Meaning}: {The} {Supervised} {Semantic} {Differential}},
	copyright = {https://creativecommons.org/licenses/by/4.0/legalcode},
	shorttitle = {Measuring {Individual} {Differences} in {Meaning}},
	url = {https://osf.io/gvrsb_v1},
	doi = {10.31234/osf.io/gvrsb_v1},
	urldate = {2026-01-02},
	publisher = {PsyArXiv},
	author = {Plisiecki, Hubert and Lenartowicz, Paweł and Pokropek, Artur and Małyska, Kinga and Flakus, Maria},
	month = nov,
	year = {2025},
}

@article{garg_word_2018,
	title = {Word embeddings quantify 100 years of gender and ethnic stereotypes},
	volume = {115},
	issn = {0027-8424, 1091-6490},
	url = {https://pnas.org/doi/full/10.1073/pnas.1720347115},
	doi = {10.1073/pnas.1720347115},
	language = {en},
	number = {16},
	urldate = {2026-01-02},
	journal = {Proceedings of the National Academy of Sciences},
	author = {Garg, Nikhil and Schiebinger, Londa and Jurafsky, Dan and Zou, James},
	month = apr,
	year = {2018},
}

@misc{perez_discovering_2022,
	title = {Discovering {Language} {Model} {Behaviors} with {Model}-{Written} {Evaluations}},
	copyright = {arXiv.org perpetual, non-exclusive license},
	url = {https://arxiv.org/abs/2212.09251},
	doi = {10.48550/ARXIV.2212.09251},
	urldate = {2026-01-02},
	publisher = {arXiv},
	author = {Perez, Ethan and Ringer, Sam and Lukošiūtė, Kamilė and Nguyen, Karina and Chen, Edwin and Heiner, Scott and Pettit, Craig and Olsson, Catherine and Kundu, Sandipan and Kadavath, Saurav and Jones, Andy and Chen, Anna and Mann, Ben and Israel, Brian and Seethor, Bryan and McKinnon, Cameron and Olah, Christopher and Yan, Da and Amodei, Daniela and Amodei, Dario and Drain, Dawn and Li, Dustin and Tran-Johnson, Eli and Khundadze, Guro and Kernion, Jackson and Landis, James and Kerr, Jamie and Mueller, Jared and Hyun, Jeeyoon and Landau, Joshua and Ndousse, Kamal and Goldberg, Landon and Lovitt, Liane and Lucas, Martin and Sellitto, Michael and Zhang, Miranda and Kingsland, Neerav and Elhage, Nelson and Joseph, Nicholas and Mercado, Noemí and DasSarma, Nova and Rausch, Oliver and Larson, Robin and McCandlish, Sam and Johnston, Scott and Kravec, Shauna and Showk, Sheer El and Lanham, Tamera and Telleen-Lawton, Timothy and Brown, Tom and Henighan, Tom and Hume, Tristan and Bai, Yuntao and Hatfield-Dodds, Zac and Clark, Jack and Bowman, Samuel R. and Askell, Amanda and Grosse, Roger and Hernandez, Danny and Ganguli, Deep and Hubinger, Evan and Schiefer, Nicholas and Kaplan, Jared},
	year = {2022},
	note = {Version Number: 1},
}

@article{zeiglerhill_narcissism_2019,
	title = {Narcissism and the pursuit of status},
	volume = {87},
	issn = {0022-3506, 1467-6494},
	url = {https://onlinelibrary.wiley.com/doi/10.1111/jopy.12392},
	doi = {10.1111/jopy.12392},
	language = {en},
	number = {2},
	urldate = {2026-01-02},
	journal = {Journal of Personality},
	author = {Zeigler‐Hill, Virgil and Vrabel, Jennifer K. and McCabe, Gillian A. and Cosby, Cheryl A. and Traeder, Caitlin K. and Hobbs, Kelsey A. and Southard, Ashton C.},
	month = apr,
	year = {2019},
	pages = {310--327},
}

@misc{mikolov_efficient_2013,
	title = {Efficient {Estimation} of {Word} {Representations} in {Vector} {Space}},
	copyright = {arXiv.org perpetual, non-exclusive license},
	url = {https://arxiv.org/abs/1301.3781},
	doi = {10.48550/ARXIV.1301.3781},
	urldate = {2026-01-02},
	publisher = {arXiv},
	author = {Mikolov, Tomas and Chen, Kai and Corrado, Greg and Dean, Jeffrey},
	year = {2013},
	note = {Version Number: 3},
}

@inproceedings{mimno_optimizing_2011,
	address = {Edinburgh, Scotland, UK.},
	title = {Optimizing {Semantic} {Coherence} in {Topic} {Models}},
	url = {https://aclanthology.org/D11-1024/},
	urldate = {2026-01-02},
	booktitle = {Proceedings of the 2011 {Conference} on {Empirical} {Methods} in {Natural} {Language} {Processing}},
	publisher = {Association for Computational Linguistics},
	author = {Mimno, David and Wallach, Hanna and Talley, Edmund and Leenders, Miriam and McCallum, Andrew},
	editor = {Barzilay, Regina and Johnson, Mark},
	month = jul,
	year = {2011},
	pages = {262--272},
}

@article{simmons_false-positive_2011,
	title = {False-{Positive} {Psychology}: {Undisclosed} {Flexibility} in {Data} {Collection} and {Analysis} {Allows} {Presenting} {Anything} as {Significant}},
	volume = {22},
	issn = {0956-7976, 1467-9280},
	shorttitle = {False-{Positive} {Psychology}},
	url = {https://journals.sagepub.com/doi/10.1177/0956797611417632},
	doi = {10.1177/0956797611417632},
	language = {en},
	number = {11},
	urldate = {2026-01-02},
	journal = {Psychological Science},
	author = {Simmons, Joseph P. and Nelson, Leif D. and Simonsohn, Uri},
	month = nov,
	year = {2011},
	pages = {1359--1366},
}

@misc{carlson_new_2025,
	title = {A {New} {Pair} of {GloVes}},
	copyright = {Creative Commons Attribution 4.0 International},
	url = {https://arxiv.org/abs/2507.18103},
	doi = {10.48550/ARXIV.2507.18103},
	urldate = {2026-01-02},
	publisher = {arXiv},
	author = {Carlson, Riley and Bauer, John and Manning, Christopher D.},
	year = {2025},
	note = {Version Number: 1},
}

@article{back_narcissistic_2013,
	title = {Narcissistic admiration and rivalry: {Disentangling} the bright and dark sides of narcissism.},
	volume = {105},
	issn = {1939-1315, 0022-3514},
	shorttitle = {Narcissistic admiration and rivalry},
	url = {https://doi.apa.org/doi/10.1037/a0034431},
	doi = {10.1037/a0034431},
	language = {en},
	number = {6},
	urldate = {2026-01-05},
	journal = {Journal of Personality and Social Psychology},
	author = {Back, Mitja D. and Küfner, Albrecht C. P. and Dufner, Michael and Gerlach, Tanja M. and Rauthmann, John F. and Denissen, Jaap J. A.},
	month = dec,
	year = {2013},
	pages = {1013--1037},
}

@misc{ines_montani_explosionspacy_2023,
	title = {explosion/{spaCy}: v3.7.2: {Fixes} for {APIs} and requirements},
	copyright = {Creative Commons Attribution 4.0 International},
	shorttitle = {explosion/{spaCy}},
	url = {https://zenodo.org/doi/10.5281/zenodo.1212303},
	urldate = {2026-01-05},
	publisher = {Zenodo},
	author = {Ines Montani and Matthew Honnibal and Adriane Boyd and Sofie Van Landeghem and Henning Peters},
	month = oct,
	year = {2023},
	doi = {10.5281/ZENODO.1212303},
}

\appendix

\section{Prolific Study Details}
\label{app:prompt}

For transparency and reproducibility, we report the exact instruction shown to participants in the AI writing task. Participants were asked to write a short text in response to the following prompt:

\begin{quote}
\small
\textbf{Prompt:} \emph{Using the box below, please write a post about artificial intelligence. It can be your opinion, a comment for an application of AI or any other thought about AI, you would like to share.}
\end{quote}

All posts in the AI corpus were produced in the English language in response to this prompt, and no additional guidance or examples were provided. Standard informed consent was obtained at the beginning of the study, and participants were informed about its aims and the nature of their participation. The study was prepared and conducted in accordance with the guidelines of the relevant psychological research ethics committee. Participants were recruited through the online crowdsourcing platform Prolific, which provides access to a diverse international pool of adult participants and allows researchers to apply demographic and eligibility pre-screening criteria. In this study, participants were recruited from the United States; the mean age was 39 years, and the gender distribution was approximately 51\% female (179), 48\% male (167), and <1\% identifying as another gender (3). All participants were members of the general Prolific participant pool and were compensated financially for their participation. Compensation was set on a task-time basis and corresponded to an average effective hourly rate of £9.10, calculated from the observed distribution of completion times in the final dataset. We report the hourly-equivalent rate to provide a transparent indicator of remuneration relative to time on task. The compensation level was consistent with Prolific’s fair-pay norms and with ethical standards for online research at the time of data collection, and was higher than the prevailing federal minimum-wage benchmark in the United States during the same period.

\section{SSD at High Dimensionality}
\label{app:high-dim}

For comparison with the PCA sweep results in the main body of the work, we reran SSD using an arbitrarily high dimensionality ($K{=}120$), corresponding to a near–full-variance reconstruction of the embedding space rather (96\% of the variance explained) than a parsimonious, stability-informed representation. All other preprocessing, regression, and clustering settings were held constant.

Tables~\ref{tab:appendix-highadm} and~\ref{tab:appendix-lowadm} report the clusters obtained at $K{=}120$ for the ADM gradient. In contrast to the sweep-selected solution, the clusters at this dimensionality exhibit lower thematic coherence and weaker alignment between lexical content and the semantic gradient, despite the SSD regression itself explaining even more of the variance than the $K{=}15$ one (adjusted $R^{2}=0.234$, $F=1.89$, $p=2.17\times10^{-5}$). Many clusters mix unrelated lexical fields (e.g., geopolitical terms, brand names, animal categories, gardening vocabulary) or aggregate language unrelated to AI discourse, indicating that the gradient begins to track idiosyncratic lexical variation rather than psychologically meaningful semantic organization. This pattern is consistent with the interpretability–stability diagnostics: when the representation is expanded to include high-variance but weakly structured components, the resulting neighborhoods become diffuse and less informative for qualitative interpretation.

\begin{center}
\small
\begin{tabular}{ccp{0.64\linewidth}}
\toprule
No. & Size & Summary (Top Words / Excerpt) \\
\midrule
1 & 51 & \emph{Mixed institutional / geographic terms}:
\textit{inclusive, asean, maldives, fiji, rolex, brics, lagos, apec} — ``AI is going to take over the world.'' \\
2 & 49 & \emph{Multilingual / function-word cluster}:
\textit{como, del, para, con, mejor, su, sistema, mundo, uso} — ``I really love AI; they are very useful to humans.'' \\
\bottomrule
\end{tabular}
\captionof{table}{High-ADM clusters at $K{=}120$.}
\label{tab:appendix-highadm}
\end{center}

\begin{center}
\small
\begin{tabular}{ccp{0.64\linewidth}}
\toprule
No. & Size & Summary (Top Words / Excerpt) \\
\midrule
1 & 27 & \emph{Animal / wildlife lexicon}:
\textit{birds, rabbits, squirrels, pigeons, sparrows, menagerie} — ``AI is going to kill jobs.'' \\
2 & 31 & \emph{Gardening / foliage terms}:
\textit{foliage, bark, potted, mulch, thyme, manure, thistle} — ``AI is a great tool for problems that have already been solved.'' \\
3 & 11 & \emph{Physical-action / nautical fragment}:
\textit{raking, hammering, trampling, keel, broadside} — ``AI just blows me away!'' \\
4 & 17 & \emph{Speculation / rumor vocabulary}:
\textit{alluding, unsubstantiated, tidbit, fable, sidenote} — ``The stupid programmers exposed their hand…'' \\
5 & 14 & \emph{Pet / illustration terms}:
\textit{terrier, labrador, postcards, ephemera} — ``Used an AI service to make some cute cat pictures.'' \\
\bottomrule
\end{tabular}
\captionof{table}{Low-ADM clusters at $K{=}120$.}
\label{tab:appendix-lowadm}
\end{center}

Overall, the $K{=}120$ solution demonstrates that maximizing variance alone leads to semantically diffuse clusters that are difficult to interpret in relation to ADM. In contrast, the sweep-selected configuration produces more coherent and psychologically meaningful gradients, underscoring the value of dimensionality choice as an interpretability-relevant decision rather than a purely statistical one.

\end{document}